\pdfoutput=1
\documentclass[11pt,a4paper]{article}
\usepackage[hyperref]{eacl2021}
\usepackage{times}
\usepackage{latexsym}

\usepackage{url}
\usepackage{stfloats}
\usepackage{amsmath}
\usepackage{amssymb}
\usepackage{graphicx}
\usepackage[export]{adjustbox}
\usepackage{subcaption}
\usepackage{footnote}
\usepackage{multirow}
\makesavenoteenv{tabular}
\makesavenoteenv{table}
\usepackage{url}
\usepackage{tikz}
\usetikzlibrary{fit,positioning}
\usepackage{booktabs}
\usepackage{tabularx}
\usepackage{makecell}
\usepackage[ruled,vlined]{algorithm2e}

\usepackage{babel} 

\DeclareMathOperator*{\KL}{KL}
\usepackage{microtype}

\aclfinalcopy 

\usepackage{xcolor}

\expandafter\def\expandafter\UrlBreaks\expandafter{\UrlBreaks
  \do\a\do\b\do\c\do\d\do\e\do\f\do\g\do\h\do\i\do\j%
  \do\k\do\l\do\m\do\n\do\o\do\p\do\q\do\r\do\s\do\t%
  \do\u\do\v\do\w\do\x\do\y\do\z\do\A\do\B\do\C\do\D%
  \do\E\do\F\do\G\do\H\do\I\do\J\do\K\do\L\do\M\do\N%
  \do\O\do\P\do\Q\do\R\do\S\do\T\do\U\do\V\do\W\do\X%
  \do\Y\do\Z}

\usepackage{soulutf8}
\usepackage[normalem]{ulem}

\usepackage{enumitem}
\setlist{nolistsep}
\usepackage{microtype}

\iffalse 
    \usepackage[final]{proofing}
  \renewcommand\hl[1]{{#1}}  
   {\draftnote{\red{#2}}}
   \newcommand\redHL[1]{}
  \newcommand\todo[1]{}

  \newcommand{\Djame}[1]{}

\newcommand{\gfcmt}[1]{}
\newcommand{\gfcorr}[2]{}
\newcommand{\jlcmt}[1]{}
\newcommand{\jlcorr}[2]{}
\newcommand{\dscmt}[1]{}
\newcommand{\dscorr}[2]{}

\else  

\usepackage[draft]{proofing}

\newcommand{\gfcmt}[1]{\textcolor{orange}{#1}}
\newcommand{\gfcorr}[2]{\textcolor{orange}{#1 $\longrightarrow$ #2}}
\newcommand{\jlcmt}[1]{\textcolor{red}{#1}}
\newcommand{\jlcorr}[2]{\textcolor{red}{#1 $\longrightarrow$ #2}}
\newcommand{\dscmt}[1]{\textcolor{brown}{#1}}
\newcommand{\dscorr}[2]{\textcolor{brown}{#1 $\longrightarrow$ #2}}

\newcommand{\Djame}[1]{
\textbf{\textcolor{red}{\hl{Djame: #1}}}
}

\newcommand\red[1]{{\textbf{\textcolor{red}{#1}}}}

\let\oldred\red
\renewcommand\red[1]{{\bf \oldred{{#1}}}}

 \newcommand\redHL[1]{\red{\hl{#1}}}
\let\olddraftnote\draftnote
\renewcommand\draftnote[1]{\olddraftnote{\red{#1}}}

\fi

\title{Disentangling semantics in language through VAEs and a certain architectural choice }

\author{Ghazi Felhi \\
  LIPN\\ Université Sorbonne Paris Nord\hspace{20px} \\ Villetaneuse, France \\
  \texttt{felhi@lipn.fr} \\\And
  Joseph Le Roux \\
  LIPN\\ Université Sorbonne Paris Nord \\ Villetaneuse, France \\
  \texttt{leroux@lipn.fr} \\\And
  Djamé Seddah \\
  INRIA Paris \\ Paris, France \\
  \texttt{djame.seddah@inria.fr} \\}

\date{}

\begin{document}
\maketitle
\begin{abstract}
We present an unsupervised method to obtain disentangled representations of sentences that single out semantic content. Using modified Transformers as building blocks, we train a Variational Autoencoder to \emph{translate} the sentence to a fixed number of hierarchically structured latent variables. We study the influence of each latent variable in generation on the dependency structure of sentences, and on the predicate structure it yields when passed through an Open Information Extraction model. Our model could separate verbs, subjects, direct objects, and prepositional objects into latent variables we identified. We show that varying the corresponding latent variables results in varying these elements in sentences, and that swapping them between couples of sentences leads to the expected partial semantic swap.
\end{abstract}
\section{Introduction}
Deep learning has brought about an insanely powerful framework to extract information from the real world through universal approximators. Consequently, a wide range of techniques have been introduced to project complex high dimensional observations such as text or images into low dimensional spaces. These low dimensional projections most often yield desirable properties such as linear separability with regard to certain high level attributes, semantically meaningful algebraic operations (\citealp{Mikolov2013}, \citealp{Bojanowski2016}) ...etc. Among these properties, disentanglement has received a lot of attention in recent studies.\\
Transparency is of great importance in the deployment of machine intelligence. In that sense, obtaining neural representation with clearly identified chunks of information is sought as a gateway to fine-grained explanation and/or controllable generation in deep learning. Interestingly, Variational Auto-Encoders (VAEs) seem to naturally disentangle neural information and have successfully been applied to this problem in numerous works (\citealp{Li2020ProgressiveRepresentations}, \citealp{John2020DisentangledTransfer}, \citealp{Chen2019ARepresentations}). This effect has been studied in depth and explained in \cite{Rolinek2019VariationalAccident}. It appears that the use of diagonal Gaussians in the VAEs' approximate posteriors, which was originally aimed at minimizing computational costs, enforces latent variables to have independent dimensions, which leads to the observed disentanglement.\\
In this work, we aim to use this property that VAEs have, and attention based language decoders to encode sentences into $N$ latent variables with similar  computational roles. These latent variables will be decoded into a sentence using co-attention, as if they represented tokens from the source language in machine translation.\\
We first relate our work to the current Natural Language Processing (NLP) landscape in section \ref{RELATED}. In sections \ref{MODEL} we describe our generative model and explain the motivation behind our design choices. Then in section \ref{OPTIM}, we construct an objective that proved effective in dealing with posterior collapse for our model. Finally (section \ref{EXPERIMENTS}), we conduct a series of experiments exhibiting quantitative and qualitative evidence that establishes our model's disentanglement capabilities.\\
Our contribution sums up as follows: We describe and architecture, and an objective that are capable of singling out semantics in a sentence without using labeled data or linguistic cues. To the best of our knowledge, we are the first to explore this research direction, and among the first few to tackle unsupervised disentanglement in NLP. As this is a first step, we will use the plain text from the SNLI dataset as was done in \cite{Schmidt2020AutoregressiveLoops} to work on homogeneous, low complexity sentences. Consequently, as opposed to mainstream language modeling studies, we will disregard long range dependencies.
\section{Related Works}
\label{RELATED}
\paragraph{Parallels between linguistics and neural architectures}
Such parallels are valuable as they enable better inductive bias in machine learning systems. RNNG \cite{Dyer2016RecurrentGrammars} and ON-LSTM \cite{Shen2019OrderedNetworks} are examples  among others (\citealp{Zhang2020Syntax-infusedGeneration}, \citealp{Du2020ExploitingApproach}, etc) of successful attempts at inducing linguistic structure in a neural language model. A plethora of \textit{post hoc} works such as (\citealp{Hu2020AModels}, \citealp{Kodner2020OverestimationModels}, \citealp{Marvin2020TargetedModels}, \citealp{Kulmizev2020DoFormalisms}) have also dived into their linguistic capabilities, the types of linguistic annotations that emerge best in them, and syntactic error analyses. The transformer based model BERT \cite{Devlin2018}, has, in turn, been subject to studies showing it operates on sentences as would a classical NLP pipeline \cite{Tenney2020BERTPipeline}, and that its attention heads perform impressively well at dependency parsing \cite{Clark2019WhatAttention}, \textit{inter alia}.  
\paragraph{Disentanglement in NLP}
As discussed by \citet{Burgess2018UnderstandingIn-VAE}, disentanglement is not only important for improving interpretability by representing high-level abstract concepts, but can also improve transfer. In contrast to the image processing field, attempts at disentanglement in NLP were mainly supervised. The main line of work revolves around multitask training schemes aimed at separating concepts in neural representations (\textit{e.g.} style vs content \cite{John2020DisentangledTransfer}, syntax vs semantics (\citealp{Bao2020}, \citealp{Chen2019ARepresentations})). A close attempt was that of \citet{Cheng2020ImprovingGuidance} which successfully disentangles a content from style using only style supervision.
\paragraph{Open Information Extraction}
Open Information Extraction (OpenIE) is the task of extracting, from a sentence, a list of predicates coupled with their arguments. The resulting tuples are handy, as they bypass complex parse trees towards a relationship-centered structure. The task can be accomplished using supervised learning on labeled samples \cite{Stanovsky2018SupervisedExtraction}, as well as earlier carefully crafted syntactic and lexical constraints \cite{Roy2020SupervisingModels}.

\section{Model}
\label{MODEL}
\subsection{Graphical Model}
\label{GRAPHMODEL}
\begin{figure}[!h]
    \centering
    \begin{minipage}[b]{0.5\textwidth}
    \begin{adjustbox}{minipage=\textwidth,scale=0.4}
    \includegraphics{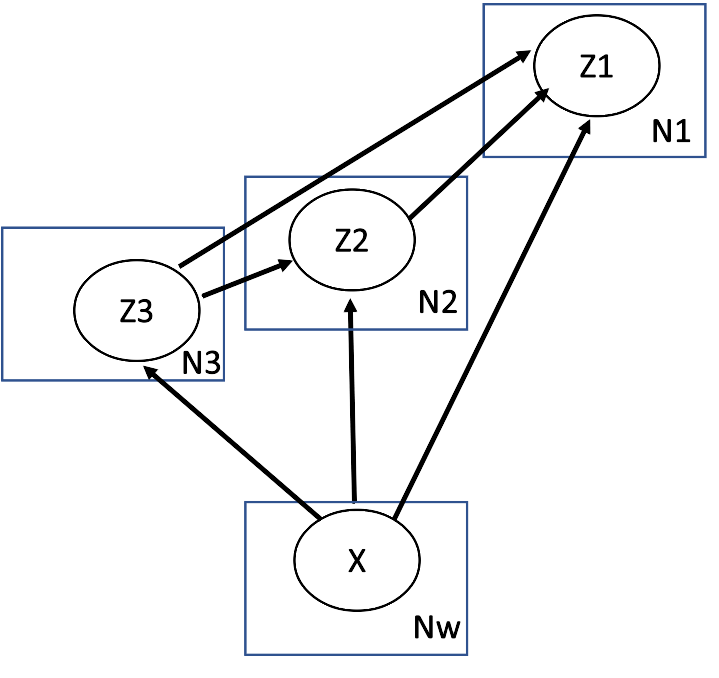}
    \end{adjustbox}
    \caption{Inference Graphical Model}
    \label{fig:INFERGRAPH}
    \end{minipage}\hfill
    \begin{minipage}[b]{0.5\textwidth}
    \begin{adjustbox}{minipage=\textwidth,scale=0.4}
    \includegraphics{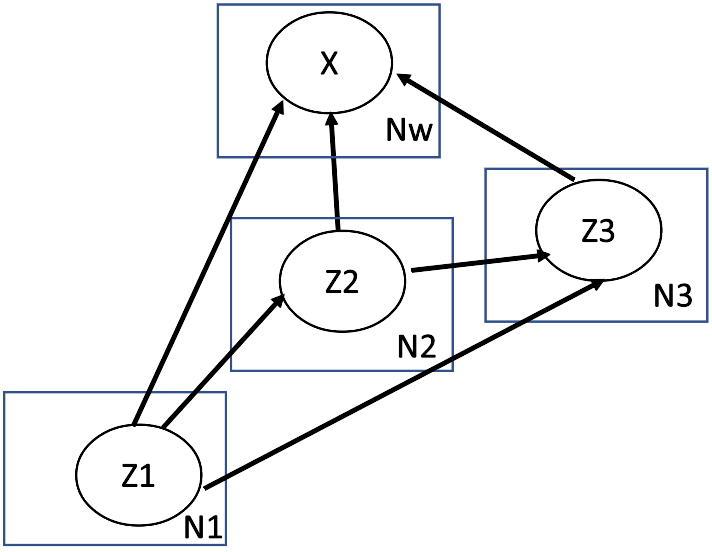}
    \end{adjustbox}
    \caption{Generative Graphical Model}
    \label{fig:GENGRAPH}
    \end{minipage}
\end{figure}

As a sentence's structure can be modeled as a tree (a dependency tree), we will make use of a hierarchy of latent variables in our model. The inference and generation graphical models are depicted in figures \ref{fig:INFERGRAPH} and \ref{fig:GENGRAPH} respectively.\\
$z_1$, $z_2$, and $z_3$ are each a set of $n_1$, $n_2$, and $n_3$ multivariate diagonal Gaussian independent latent variables of size $z_{size}$. A fixed standard normal distribution $p$ is set as a prior for $z_1$ in the generative model. Consequently, the generative model decomposes into $p_\theta(x, z_1, z_2, z_3) = p(z1)p_\theta(z_2|z_1)p_\theta(z_3|z_1, z_2)p_\theta(x|z_1, z_2, z_3)$, and the inference model decomposes into $p_\phi(x, z_1, z_2, z_3)=p_{data}(x)p_\phi(z_3|x)p_\phi(z_2|z_3, x)p_\phi(z_1|z_2, z_3, x)$, so that $p_{data}$ is the true data distribution.\\
The neural components modelling the different conditional distributions hereabove will be described in the upcoming sections.

\subsection{Encoder}
\begin{figure}[!h]
    \begin{adjustbox}{minipage=\textwidth,scale=0.5}
    \includegraphics[width=\textwidth]{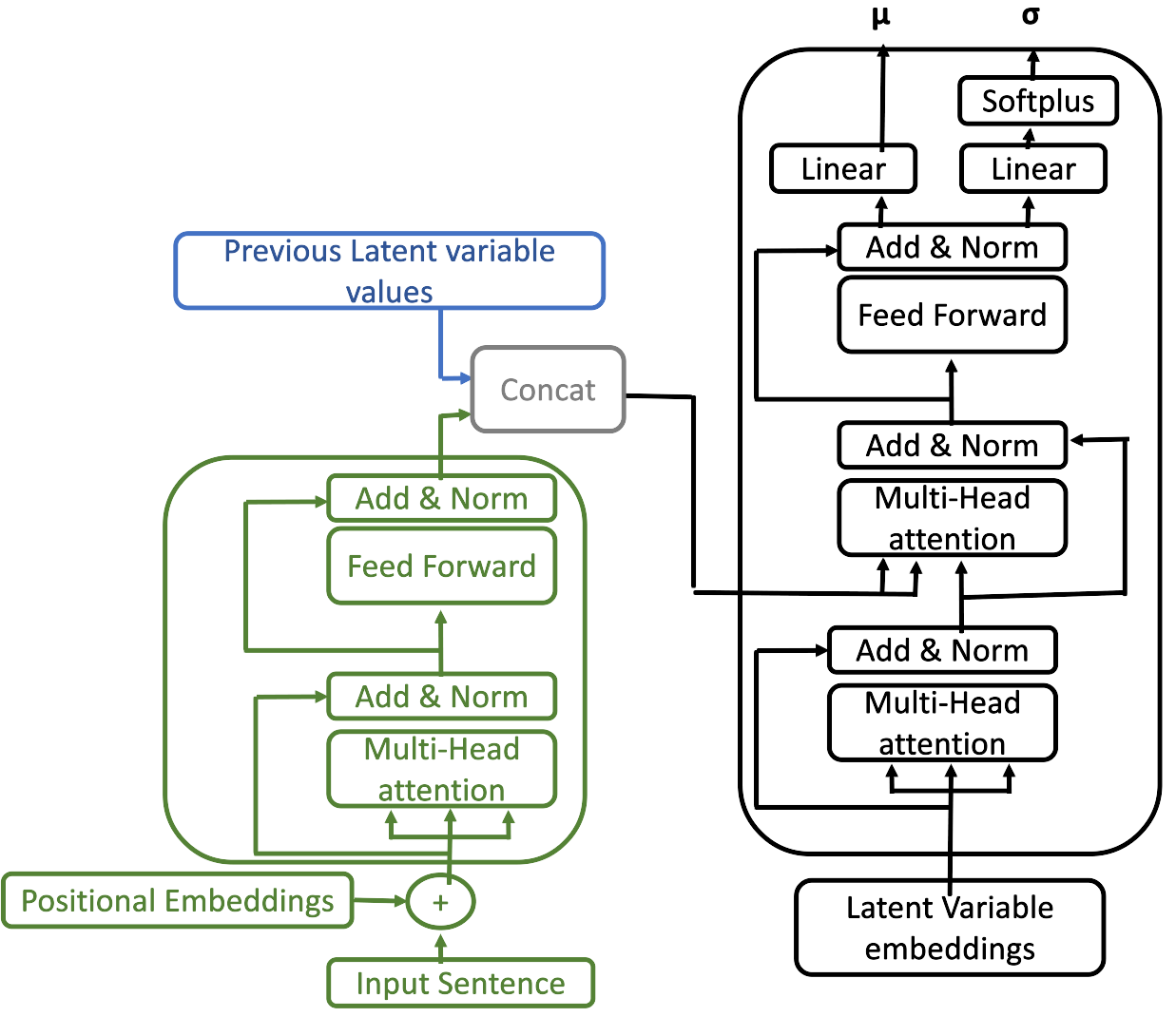}
    \end{adjustbox}
    \caption{The Encoder}
    \label{fig:TransEnc}
\end{figure}

\paragraph{Constructing $p_\phi(z_3|x)$: }
The model differs from classical VAE encoders in that it will encode a sentence into $n_3$ latent variables, where $n_3$ is a fixed integer (\emph{regardless} of the sentence length).\\
Our choice was to use a transformer \cite{Vaswani2017}. More specifically, we will use the transformer encoder-decoder architecture that is mostly used for machine translation. Contrary to an encoder only transformer , this architecture allows for obtaining a number of output elements that is different from that of the input sequence (as is needed for translation).  It has been established that transformers can store sentence-level statistics in artificially introduced tokens (\textit{e.g.}\emph{SEP} in \cite{Devlin2018}). In a similar manner, we will feed a set of fixed $n_3$ learnable vectors to the decoder in place of it s targets. The transformer encoder-decoder's architecture and the decoding process are explicited in figure \ref{fig:TransEnc}, where the "Previous Latent variable value" placeholder (light blue) is empty.We will apply $n_3$ distinct linear transformations (resp. $n_3$ MLPs with softplus activations) to obtain the means (resp. the standard deviations) of the posterior distribution of $z_3$.

\paragraph{Constructing $p_\phi(z_2|z_3, x)$ and $p_\phi(z_1|z_2, z_3, x)$: }
Similarly to the way we obtain $z_1$, we use a Transformer encoder-decoder architecture. The latent variables that we condition on are introduced here by concatenating them to the input sentence after positional-encoding and Transformer-encoding, as depicted in figure \ref{fig:TransEnc}. These latent variables are viewed as additional elements of the sentence with no specific positioning.

\begin{figure}[!h]
    \begin{adjustbox}{minipage=\textwidth,scale=0.5}
    \includegraphics[width=\textwidth]{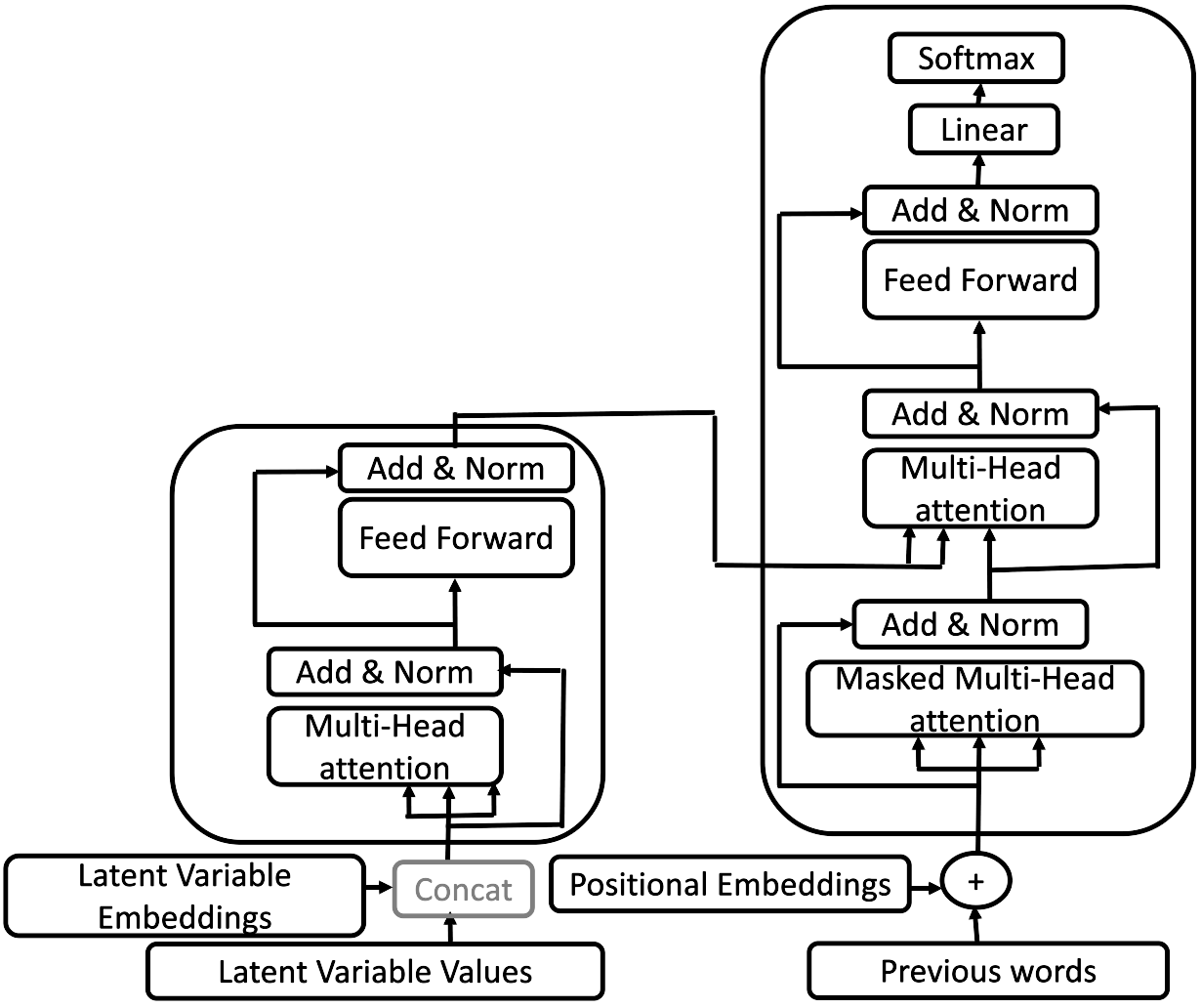}
    \end{adjustbox}
    \caption{The Decoder}
    \label{fig:TransDec}
\end{figure}

\subsection{Decoder}

\paragraph{Constructing $p_\theta(z_2|z_1)$ and $p_\theta(z_3|z_1, z_2)$: }
As explained in \ref{GRAPHMODEL}, we use a learnable structured prior $p_\theta(z_1, z_2, z_3) = p(z1)p_\theta(z_2|z_1)p_\theta(z_3|z_1, z_2)$. To obtain the parameters of $z_2$ and $z_3$ in the generative model, we use the same architecture we used in the encoder without inputting text (\textit{i.e} we use the model in figure \ref{fig:TransEnc} while dropping the green part). 
\paragraph{Latent Variable Identifier}
One must notice that, given our training procedure (\textit{c.f.} ELBo in section \ref{OPTIM}), all latent variables in $z_1$ are enforced to follow the same prior. In the generation step, we will be sampling a set of similarly distributed random variables with no means for the decoding network to distinguish between them. As our objective is to encode $n_1$ different types of information in $z_1$, and to have the decoder identify and leverage this information, we will concatenate the vectors corresponding to the value of each latent variable in $z_1$ to a latent-variable-specific trainable vector before having it decoded. The same will be done for $z_2$, and $z_3$ even though their trainable priors enable better distinguishability .
\paragraph{Sequence decoder $p_\theta(x|z_1, z_2, z_3)$}
Sequence to sequence models (Seq2seq \cite{Sutskever2014}) that do not use attention were always found to be lacking in comparison to those that do. As a side effect to our architectural choices, we will be able to use attention based decoders and thus benefit from their higher expressiveness. In the same spirit as that of the previous section, we will use sequence transduction components that were originally designed to be used in machine translation to simultaneously translate and align.\\
We chose to use here the same transformer encoder-decoder architecture used in the encoding stage, but with different inputs. It will closely follow machine translation in this step by receiving the latent variable values as source inputs, and the previously generated tokens as target inputs. Contrary to what is done in the sequence encoder, the transformer applied to targets will use an attention mask that enforces the current generated word to depend only on previous words. A Latent variable identifier coupled with a transformer decoder are depicted in figure \ref{fig:TransDec}.
\section{Optimization}
\label{OPTIM}
Preliminary experiments have revealed that this model was subject to severe posterior collapse. Using $\KL$-annealing \cite{Bowman2016GeneratingSpace}, or its combination with $\KL$-thresholding \cite{Li2020AText} was not effective in yielding adequate results. As $\KL$-thresholding forces all the latent dimensions to stay at least $\gamma$ bits away from the prior ($\gamma$ being the threshold), it may create artificial redundancy in the latent variables, which counteracts disentanglement.\\ 
In the following, we will describe a procedure that turned out to bring satisfactory generation results while keeping this generation dependent on the latent variables.\\
The original objective of VAEs is the Evidence Lower Bound (ELBo):
\begin{multline}
    \log p_\theta(x) \geq\\ \mathbb{E}_{(z_1, z_2, z_3) \sim q_\phi(z_1, z_2, z_3|x)}\left[ \log p_\theta(x|z_1, z_2, z_3) \right] -\\
    \KL[q_\phi(z_1, z_2, z_3|x)||p_\theta(z_1, z_2, z_3)]
\end{multline}
Where $\KL[.]$ is the Kullback-Leibler divergence. The first term of the right hand side is the reconstruction term. The second term represents the information we get about our latent variables from the observation $x$. "Posterior collapse" happens when this term collapses to zero (\textit{i.e.} when $x$ brings no more information on $z1$, $z2$, and $z3$ than what was described by the prior). The upcoming alternative objective aims at keeping this term to a multiple of the reconstruction's value, while spreading the information gain from the observation across the 3 levels of latent variables:
\begin{multline}
    \max ( \mathbb{E}_{(z_1, z_2, z_3) \sim q_\phi(z_1, z_2, z_3|x)}\left[ \log p_\theta(x|z_1, z_2, z_3) \right] ,\\
     - \alpha \beta {\KL}_{max})\\
    s.t. \hspace{5 px} {\KL}_{max}=
    \max(\\\mathbb{E}_{(z_2, z_3) \sim q_\phi(z_2, z_3|x)}\left[\KL[q_\phi(z_1|z_2, z_3, x)||p(z_1)] \right],\\
    \mathbb{E}_{(z_1, z_3) \sim q_\phi(z_1, z_3|x)}\left[\KL[q_\phi(z_2|z_3, x)||p_\theta(z_2|z_1)] \right],\\
    \mathbb{E}_{(z_1, z_2) \sim q_\phi(z_1, z_2|x)}\left[\KL[q_\phi(z_3|x)||p_\theta(z_3|z_1, z_2)] \right])
\end{multline}
The global max ensures that we are minimizing the selected Kullback-Leibler divergence up to $\frac{1}{\alpha\beta}$ times the reconstruction loss so far. The values of $\alpha$ and $\beta$ will be discussed in section \ref{SETUP}. In contrast to $\KL$-thresholding, this objective thresholds each latent variable layer \emph{as a whole}, and uses a \emph{mobile} threshold that is linear in the reconstruction loss of the example at hand. A lower value of $\alpha\beta$ allows for a better perplexity at the cost of a lower $\KL$-divergence (more posterior collapse), while a higher value guarantees more informative posteriors at the cost of a higher perplexity (empirically leading to semantically inconsistent sentences). As for $\KL_{max}$, it ensures that we are optimizing the hierarchy level that strays most from the prior for each example. In fact, when using structured generative models, the first layer tends to absorb all the mutual information with observations while the subsequent layers are hardly informative about the observation. This behavior was demonstrated and studied in depth by \cite{Zhao2017LearningModels}, and confirmed by our preliminary experiments. \\

\begin{figure*}[!h]
    \hskip -5px
    \begin{adjustbox}{minipage=\textwidth,scale=1.0}
    \includegraphics[width=\textwidth]{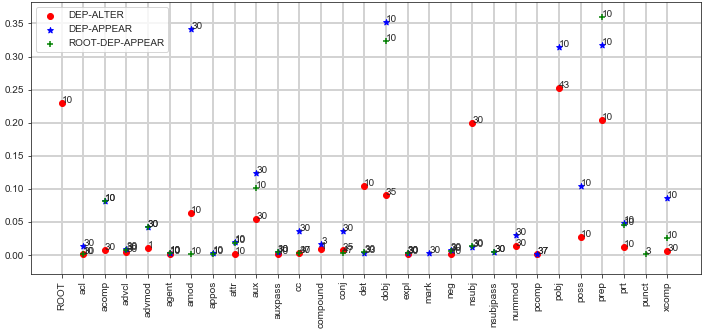}
    \end{adjustbox}
    \caption{Latent variables that influence the most each of our dependency parsing metrics for each dependency label. The vertical position of each point is the probability of influencing the metric. }
    \label{fig:QUANTDEP}
\end{figure*}
\section{Experiments}
\label{EXPERIMENTS}
\subsection{Setup}
\label{SETUP}
As previously mentioned, our training set consists of low complexity text extracted for the SNLI dataset by \citet{Schmidt2020AutoregressiveLoops}. The sentences are on average $8.92\pm 2.66$ tokens long. We use 90K samples as a training set, and 10K samples as a test set.\\
We found it best for disentanglement to train the model with more latent variables than it needs, instead of fixing the number of latent variables to the expected number of disentangled concepts. This observation is not surprising, as it is well known that overparametrized neural networks have higher chances of containing well initialized subnetworks \cite{Frankle2018TheNetworks}. $n1$, $n2$, and $n3$ are therefore fixed to 16 each. Training details are in Appendix \ref{TRAININGAPPEN}.The code for training our model, and performing the evaluations below is publicly available\footnote{\href{https://github.com/ghazi-f/Disentanglement\_Transformer}{https://github.com/ghazi-f/Disentanglement\_Transformer}}.
\subsection{Evaluation Protocol}
We analyze our models qualitatively as well as quantitatively. Our quantitative analysis partly relies on the OpenIE system of \citet{Stanovsky2018SupervisedExtraction}\footnote{Online Live demo from AllenNLP \href{https://demo.allennlp.org/open-information-extraction}{https://demo.allennlp.org/open-information-extraction}}. We obtain the necessary statistics  with the following process:\\
We samples 100 sentences from the model. Then, for each sentence, we resample 10 times the 48 latent variables one at a time and generate the resulting new sentence. This results in 48K (original sentence, modified sentence) couples. After parsing all the sentences using \cite{spacy2}, and obtaining their first\footnote{The predicates that follow the first OpenIE predicate correspond to subordinates clauses} predicate structure using \cite{Stanovsky2018SupervisedExtraction}, we calculate the following between the original and the modified sentences:
\begin{enumerate}
    \item \textit{ROOT-DEP-APPEAR:} the set of non-common dependency labels in the children of ROOT.
    \item \textit{DEP-APPEAR:} the set of non-common dependency labels over the whole sentence.
    \item \textit{OIE-APPEAR:} the set of non-common OpenIE labels over the whole sentence.
    \item \textit{DEP-ALTER:} if both sentences have the same length, we extract the list of dependency labels for which the text spans have changed.
    \item \textit{OIE-ALTER:} if both sentences have the same first predicate structure, we extract the list of predicate arguments for which the text spans have changed.
\end{enumerate}
This information is used to calculate statistics about the influence of each latent variable in the model on the generated sentence. *-APPEAR variables (resp. *-ALTER variables) are used to analyze the influence of latent variables on the structure (resp. content) of sentences.

\begin{figure*}[!h]
    \hspace{-20px}
    \begin{adjustbox}{minipage=\textwidth,scale=1.1}
    \includegraphics[width=\textwidth]{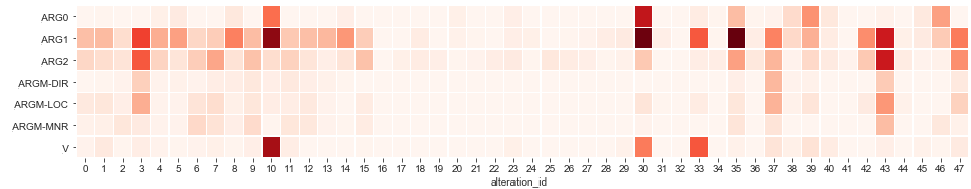}
    \end{adjustbox}
    \caption{The influence of each latent variable on the \textbf{content} of OpenIE arguments. The scale goes from light (low) to dark (high). }
    \label{fig:RELDIFF}
\end{figure*}

\subsection{Quantitative results}
\paragraph{From a dependency structure perspective}
\textit{ROOT-DEP-APPEAR}, \textit{DEP-APPEAR}, and \textit{DEP-ALTER} are lists of dependency labels. We found it interesting to look at the latent variables that causes each of the dependency labels to appear in each of the 3 lists.\\
Influencing \textit{ROOT-DEP-APPEAR} means having the corresponding dependency label appear/disappear from the ROOT children. Influencing \textit{DEP-APPEAR} means having the corresponding dependency label appear/disappear from the whole dependency tree. Influencing \textit{DEP-ALTER} means having the text behind a certain dependency label change in a static length sentence.\\
We report the latent variables with the highest influence on each dependency label for these three statistics in figure \ref{fig:QUANTDEP}. A first look at the results shows that a major part of the variability in the generated sentences is expressed by latent variable (LV) 10. Figure \ref{fig:QUANTDEP} shows that is it responsible for the content of the ROOT node, which explains how its influence propagates to the major part of the sentence. LV 30 seems to influence nominal subjects (passive or active), auxiliaries (possibly for conjugation) in terms of content, and  numeral modifiers, and expletives when it comes to structure. These cues clearly point to LV 30 being responsible for subject related information. Other highly influential LVs are 35, and 43. 35 is responsible for the appearance of conjunctions, as well as the content of direct objects. LV 43 controls the content in prepositional objects, and is structurally related to the appearance of compounds, adverbial clause modifiers, and markers. Consequently, we expect these last two LVs to control most of the information past verbs and subjects.

\paragraph{From an OpenIE perspective}
We plot the influence of each latent variable on the appearance of OpenIE arguments (\textit{OIE-APPEAR}) and on their content (\textit{OIE-ALTER}) as heat maps. \textit{OIE-APPEAR} showed no evidence of the presence of variables that control structure while disregarding content. Therefore the heat map for \textit{OIE-APPEAR} is reported in Appendix \ref{OIE-APPEAR-APPENDIX}, while the heat map for \textit{OIE-ALTER} is in figure \ref{fig:RELDIFF}.\\
As was expected from the dependency parsing analysis, LV 10 is the most influential on the verb. It can also be seen that LV 30 has the highest influence on ARG0 (\textit{i.e.} the subject).  Along with the information from the dependency analysis, figure \ref{fig:RELDIFF} further stresses the roles of LV 35 and LV 43. LV 35 seems to specialize in the direct object (ARG1), while LV 43 partly describes the direct object as well as secondary arguments (ARG2 often corresponds to prepositional objects), and contextual arguments (ARGM-DIR, ARGM-LOC, and ARGM-MNR correspond to direction, location, and manner). One should notice that LV 10 is in the root level of latent variables ($10<16$), LV 30 in the middle level ($16<30<32$) and LVs 35 and 43 in the leaf level ($32<35<43<48$). The disentangled information is consequently also arranged as dictated by a linguistic dependency structure. This further confirms the ability of machine learning models to align with our conception of linguistic structure.\\
\subsection{Qualitative results}
Here we will exhibit some samples where the latent variables were varied in different manners. We will take a special interest in LVs 10, 30, 35, and 43 as these have shown potential for interpretability. Table \ref{tab:resultsresample} shows an example where we altered a latent variable for some sentences.\\
A second experiment we did, was to swap the value of certain latent variables between two sentences. The results are in table \ref{tab:resultsswap}.
\paragraph{LV 10} As was pointed out by the quantitative analyses, this LV is the most influential (overall) on the sentence. But it seems to specialize, to a certain extent, in specifying the verb. In table \ref{tab:resultsresample}, We can see that varying LV 10 keeps the same subject for the sentence, but varies the verb and the object (which is highly dependent on the verb). As LV 10 is a low level variable, changing it results in an incompatibility with the higher level variables, and a radical change in the sentence is observed. The fact that verbs can't be changed independently from their subject is also observed in table \ref{tab:resultsswap}. In fact, swapping LV 10 clearly results in unexpected changes.

\begin{table*}[t]
    \small
    \centering
    \begin{tabularx}{16cm}{|X|c|X|X|X|}
    \hline
     Original sentence&ALV& Sample 1& Sample 2 & Sample 3 \\
    \hline \hline
     a girl is holding a ball &10& a girl is riding a bike down the sidewalk & a girl is making a toy & a girl is sitting at a table
    \\\hline
     a child is running in the park &10& a child is in a store & a child is standing on a bench  & a child is painting a marathon\\\hline
     a man is looking at something &10& a man is playing with his dog & a man is cooking in a park & a man is in a dress \\\hline
     a girl is holding a ball &30& a man is holding a ball & a group of people are sitting down &a kid is holding a ball
     \\\hline
     a child is running in the park &30& two men are running in the street & a boy is running in the park& a man is looking at a large boat .\\\hline
    a man and a woman are sitting in a race &30& a man is laying on a bench & a child is laying on a bench &  a kid is laying on a bench\\\hline
     a girl is holding a ball &35& a girl is holding a bicycle & a girl is holding a baseball & a girl is holding a baby \\\hline
     two girls are wearing a pink and pink shirt &35& two girls are wearing a hat and talking about to get to get  & two girls are wearing a red and pink shirt  & two girls are wearing a green and pink shirt  \\\hline
     a group of people are standing in a park &35& a group of people are dancing in a park & a group of people are in a parade &  a group of people are walking in a park \\\hline
      two girls are wearing a pink and pink shirt &43& two girls are wearing a pink uniform & two girls are wearing a pink dress &two girls are wearing a pink hat \\\hline
       a man is sitting in a chair &43&  a man is sitting in a $<?>$ & a man is sitting in a park & man is sitting in a house\\\hline
        a group of people are sitting around a table &43& a group of people are sitting at a table & a group of people are sitting in a restaurant & a group of people are sitting on a beach \\\hline
     \end{tabularx}
    \caption{Varying the value of a specific latent variable for a sentence. ALV is the Altered LV.}
    \label{tab:resultsresample}
\end{table*} 

\begin{table*}[t]
    \small
    \centering
    \begin{tabularx}{16cm}{|X|X|c|X|X|}
    \hline
     Sentence 1& Sentence 2& SLV& Swapped Sentence 1 & Swapped Sentence 2 \\
    \hline \hline
      two people are outside at a store & a man is wearing a helmet &30& a man is outside with a red shirt & two people are wearing a helmet \\\hline
      a child is jumping in the snow & a boy is running through the snow &30& a boy is jumping in the snow & a child is running through the snow \\\hline
      a boy is using a phone & a man is walking on a sidewalk &30& a man is using a phone & a boy is walking on a sidewalk \\\hline
      a young boy is playing with a ball & a little girl jumps on a bike &10& a man is singing in a park & a little girl is playing in a park \\\hline
      a person is looking at a boat & a young boy is playing with a ball &10& a person is playing with a ball & a man is riding a horse \\\hline
       a little girl jumps on a bike & a man is taking a nap &10& a little girl is running in the water & a man is standing on a bench \\\hline
        a person is riding a bicycle on a beach & a man is riding a bike &35& a person is riding a bike on a beach & a man is riding a bicycle \\\hline
         a couple of people are playing with a little girl & a snowboarder is playing with a white mountain &43& a couple of people are playing in a city & a snowboarder is playing with a child \\\hline
         a boy is holding a ball & a man is holding a sign &35& a boy is holding a sign & a man is holding a ball \\\hline
      \end{tabularx}
    \caption{Swapping the value of a specific latent variable between two sentences. SLV is the swapped LV. }
    \label{tab:resultsswap}
\end{table*} 
\paragraph{LV 30} Despite some negative examples(table \ref{tab:resultsresample}, 5th row, 5th column), Tables \ref{tab:resultsresample} and \ref{tab:resultsswap} clearly demonstrates that LV 30 contains the information on the subject. We can see, nevertheless, that changing it results in co-adaptation of the rest of the sentence, such as the conjugation of a verb (table \ref{tab:resultsresample}, 5th row, 3rd column). A surprising observation can be made in table \ref{tab:resultsswap}, 6th row: a change of subject from plural to singular resulted in the same co-adaptation of the verb on 3 examples. It is unclear whether "sitting" has been reinterpreted as "laying", or the latent code stores the action for a group in a different area than the action for a single individual.
\paragraph{LVs 43 an 35} These two LVs, most often encode low level information (\textit{i.e.} leaf information in the dependency parsing sense). To generate table \ref{tab:resultsswap}, we had to try both LVs and see which contained the information for the verb at hand. Another constraint for results in table \ref{tab:resultsswap} to be coherent, was for the sentences to feature the same verb. As these LVs are leaf LVs, it is only natural that they can only be swapped between sentences where they remain a high probability sample when conditioned on the root LVs.
 As can be seen in table \ref{tab:resultsresample} (rows 8 and 10) when varying these LVs for the same sentence, they change different aspects of the object. We could also confirm that LV 35 most often controls direct objects (table \ref{tab:resultsresample} lines 7, 8, and table \ref{tab:resultsswap} lines 7, and 9), while LV 43 holds the information on prepositional objects (table \ref{tab:resultsresample} lines 11 and 12, and table \ref{tab:resultsswap} line 8). LV 35 also seems to control some \emph{intransitive} verbs (table \ref{tab:resultsresample} line 9). Given that most sentences are in the past, these verbs may be perceived by the model as objects to the auxiliary. 
 
 \paragraph{Encoder-Decoder Discrepancy}
An inherent short-coming of VAEs is the fact their objective (ELBo) is only a lower bound to the exact marginal log-likelihood of the data. In fact, the positive term quantifying the gap between ELBo and $log(x)$ is $KL[q(z|x)||p(x|z)]$. This difference results in a discrepancy between encoder and decoder. We study this discrepancy under the light of the attention values between input tokens, and latent variables. Example heat maps of these attention values for our latent variables of interest (10, 30, 35, and 43) are provided in Appendix \ref{ATTENTION-APPENDIX}. The most striking observation that can be made is that our latent variables attend to positions where the information they need is expected to appear, with little reliance on the tokens present in the attended positions. In fact, LV 10, LV 30, LV 43, and LV 35 almost exclusively attend to positions $ \left[3-5\right] $, $ \left[1-3\right] $, $ \left[5-6\right] $, and $ \left[6-7\right]$. Fortunately, the latent variables in the generator do not seem to only influence the same predefined positions as is illustrated by the examples in tables \ref{tab:resultsresample} and \ref{tab:resultsswap} (\textit{e.g.} LV 43 influences tokens at position 9 in the last example of table \ref{tab:resultsresample}). A secondary observation that can be made is that LV 10 and LV 30 attend a lot to latent variables from previous layers. This establishes that the encoder is actively using its latent variable structure (\textit{i.e} it successfully learned a structured posterior). 
 \section{Discussion \& Conclusion}
 Our model's capabilities differ from unsupervised OpenIE in that it factors information that aligns with predicate arguments instead of extracting text spans that correspond to these arguments. This is demonstrated by the fact that some words have information from more than one of our disentangled latent variables (direct objects are defined both with information from LV 10 and LV 35).
 It is also noteworthy that our model is limited with regard to two aspects. The first is that we could not discover structure related disentangled information in its latent variables. In that regard, future iterations may be able to obtain improvements through a fine grained use of self attention (separate latent variables for keys, queries, and values), or through non-sequential generation.  The second limit is the posterior collapse, which was handled to a certain extent by our modified ELBo. By calibrating $\alpha\beta$ we could compromise between low perplexity and high $\KL$-divergence, but a great proportion of the generative model's descriptive capacity still resides in $p_\theta(x|z_1, z_2, z_3)$ instead of $p_\theta(z_1, z_2, z_3)$. In fact, contrary to our expectations, sequential sampling from $p_\theta(x|z_1, z_2, z_3)$ with fixed $(z_1, z_2, z_3)$ (as opposed to greedy sampling) did not yield paraphrases with fixed semantics. Our model is therefore expected to greatly improve with future strives in dealing with posterior collapse. 
 To the best of our knowledge, our model is the first to induce a form of disentanglement that separates semantics in a sentence. Through our analysis, we could highlight 4 latent variables with distinguishable semantic content.  Moreover, this model is also the first to accomplish disentanglement on language without any form of supervision (\textit{e.g} labels, linguistic cues, etc).  Hence, it's inductive bias could serve as a basis to derive semi-supervised models, weakly supervised models or other forms of data efficient machine learning models. Data efficiency is central to contemporary NLP as annotated data is getting more expensive with the explosive rise of User Generated Content and the concomitant annotation difficulties \cite{Seddah2020BuildingHell}. A highly potent research direction and a natural extension of our work would be to explore the results of applying our method to word level representations (disentangling morphological phenomena) and to document level representations (disentangling rhetorical structure).

\section*{Acknowledgments}
We want to thank Antoine Simoulin for his proofreading and valuable comments. This work is supported by the PARSITI project grant (ANR-16-CE33-0021) given by the French National Research Agency (ANR), the \emph{Laboratoire d’excellence “Empirical Foundations of Linguistics”} (ANR-10-LABX-0083), as well as the ONTORULE project. It was also granted access to the HPC resources of IDRIS under the allocation 20XX-AD011012112 made by GENCI.

\bibliography{references, bib2}
\bibliographystyle{acl_natbib}
\clearpage
\appendix
\section{Training details}
\label{TRAININGAPPEN}
 We use a 48-dimensional Transformers with 4 attention heads, 2 layers for each of the encoder modules, and 3 layers for each of the decoder modules. The model is warmed up by training it in pure reconstruction ($\alpha=0$) for 3000 steps, then annealing the $\KL$-divergence (linearly raising $\alpha$ to 1) during 3000 steps. $\beta$ is initialized at 6, then decreased by 1 each time the perplexity (evaluated each 3 epochs) stops decreasing. The training is halted when $\beta$ reaches 3 and the  perplexity stops decreasing. This setup has been reached through a manual search of the hyper parameters that best exhibited the behavior we sought. During evaluation, we sample sentences (conditioned on the latent variables we sampled beforehand) in a greedy fashion.
\begin{figure*}[!t]
    \hspace{-18px}
    \begin{adjustbox}{minipage=\textwidth,scale=1.05}
    \includegraphics[width=\textwidth]{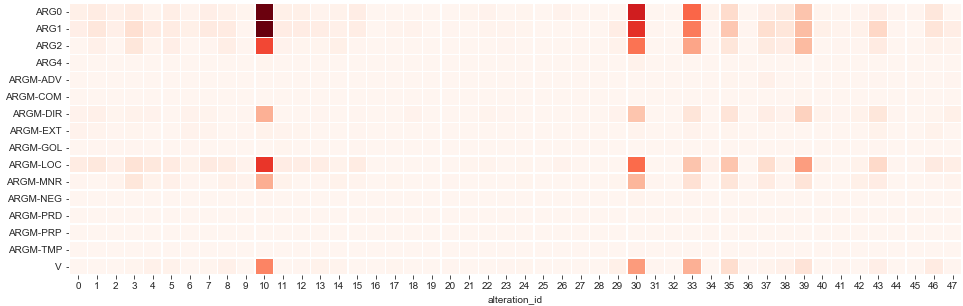}
    \end{adjustbox}
    \caption{The influence of each latent variable on the \textbf{appearance} of OpenIE arguments. The scale goes from light (low) to dark (high). }
    \label{fig:NEWRELS}
\end{figure*}
\section{OIE-APPEAR}
\label{OIE-APPEAR-APPENDIX}
The heat map for OIE-APPEAR is plotted in figure \ref{fig:NEWRELS}.
\section{Encoder Attention}
\label{ATTENTION-APPENDIX}
\begin{figure*}[h!]
    \begin{adjustbox}{minipage=\textwidth,scale=1.2}
    \hspace{-60px}
    \adjincludegraphics[width=\textwidth, trim={0 0 {.25\width} 0},clip]{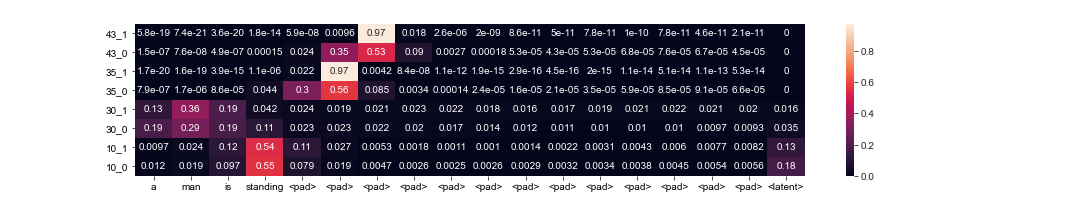}
    \end{adjustbox}
    \begin{adjustbox}{minipage=\textwidth,scale=1.2}
    \hspace{-60px}
    \adjincludegraphics[width=\textwidth, trim={0 0 {.25\width} 0},clip]{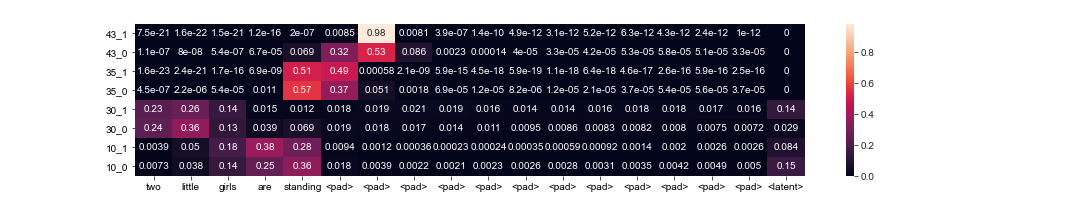}
    \end{adjustbox}
    \begin{adjustbox}{minipage=\textwidth,scale=1.2}
    \hspace{-60px}
    \adjincludegraphics[width=\textwidth, trim={0 0 {.25\width} 0},clip]{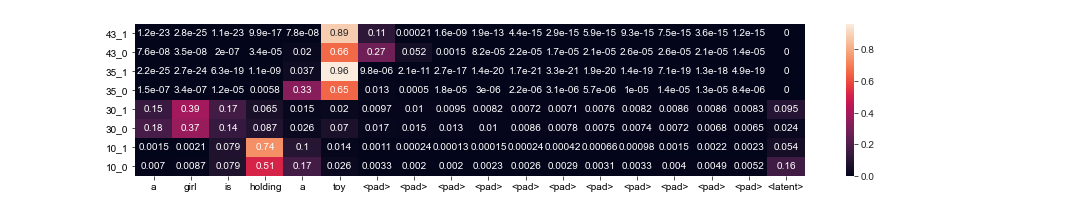}
    \end{adjustbox}
    \begin{adjustbox}{minipage=\textwidth,scale=1.2}
    \hspace{-60px}
    \adjincludegraphics[width=\textwidth, trim={0 0 {.25\width} 0},clip]{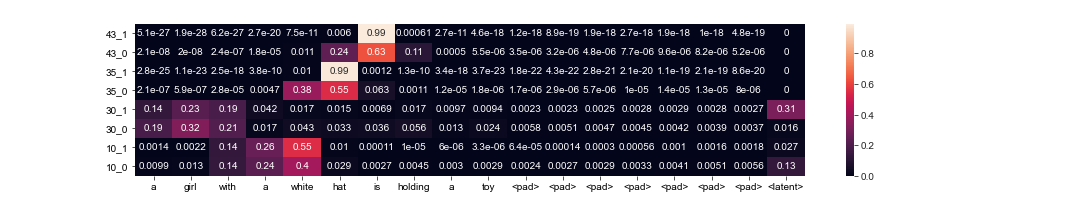}
    \end{adjustbox}
    \caption{First 4 encoder attention examples. The y-axis labels are to be read "$<$latent variable index$>\_<$encoder layer$>$". The lighter the color of the box, the higher the attention value. The last column $<$latent$>$ is the summation of the attention values between the indicated latent variable, and the latent variables from the previous latent variable layer (\textit{c.f.} figure \ref{fig:INFERGRAPH} for the encoder latent variable structure). }
    \label{fig:att1}
\end{figure*}

\begin{figure*}[h!]
    \begin{adjustbox}{minipage=\textwidth,scale=1.2}
    \hspace{-60px}
    \adjincludegraphics[width=\textwidth, trim={0 0 {.25\width} 0},clip]{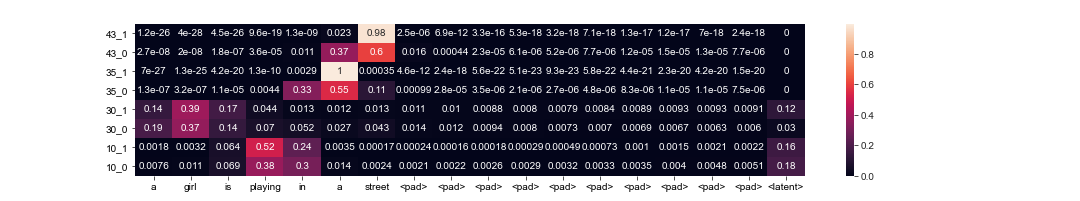}
    \end{adjustbox}
    \begin{adjustbox}{minipage=\textwidth,scale=1.2}
    \hspace{-60px}
    \adjincludegraphics[width=\textwidth, trim={0 0 {.25\width} 0},clip]{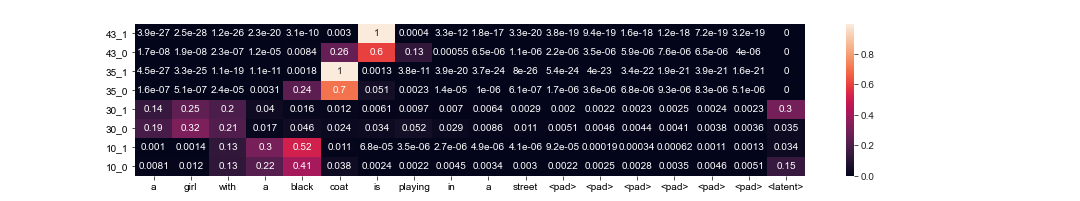}
    \end{adjustbox}
    \begin{adjustbox}{minipage=\textwidth,scale=1.2}
    \hspace{-60px}
    \adjincludegraphics[width=\textwidth, trim={0 0 {.25\width} 0},clip]{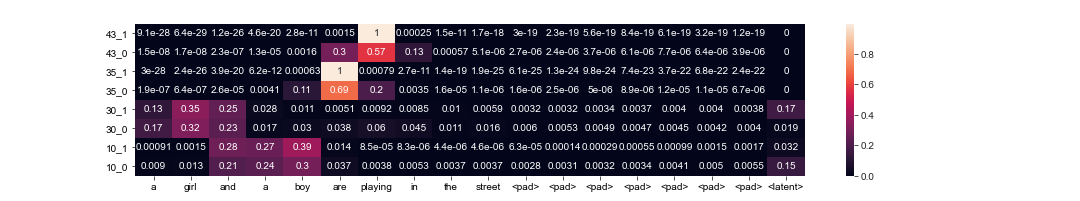}
    \end{adjustbox}
    \begin{adjustbox}{minipage=\textwidth,scale=1.2}
    \hspace{-60px}
    \adjincludegraphics[width=\textwidth, trim={0 0 {.25\width} 0},clip]{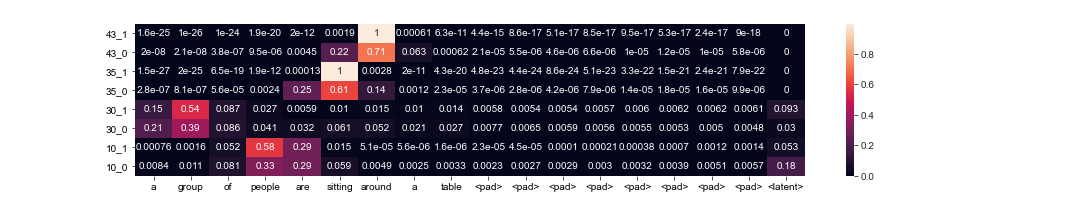}
    \end{adjustbox}
    \caption{Second 4 encoder attention examples, generated in the same way as those of figure \ref{fig:att1}}
    \label{fig:att2}
\end{figure*}

We provide the attention heat maps illustrating our qualitative attention analysis in figures \ref{fig:att1} and \ref{fig:att2}.

\end{document}